\pdfoutput=1

\documentclass[11pt]{article}

\usepackage[final]{acl}

\usepackage{times}
\usepackage{latexsym}

\usepackage[T1]{fontenc}

\usepackage[utf8]{inputenc}

\usepackage{microtype}

\usepackage{inconsolata}

\usepackage{graphicx}
\usepackage{float}
\usepackage{hyperref}
\usepackage{subcaption}

\title{Efficient Machine Translation Corpus Generation: Integrating Human-in-the-Loop Post-Editing with Large Language Models}

\author{Kamer Ali Yuksel, Ahmet Gunduz, Abdul Baseet Anees, Hassan Sawaf \\
        aiXplain Inc., Los Gatos, CA, USA \\
        \{kamer, ahmet, abdul.anees, hassan\}@aixplain.com}

\begin{document}
\maketitle
\begin{abstract}
This paper introduces an advanced methodology for machine translation (MT) corpus generation, integrating semi-automated, human-in-the-loop post-editing with large language models (LLMs) to enhance efficiency and translation quality. Building upon previous work that utilized real-time training of a custom MT quality estimation metric, this system incorporates novel LLM features such as Enhanced Translation Synthesis and Assisted Annotation Analysis, which improve initial translation hypotheses and quality assessments, respectively. Additionally, the system employs LLM-Driven Pseudo Labeling and a Translation Recommendation System to reduce human annotator workload in specific contexts. These improvements not only retain the original benefits of cost reduction and enhanced post-edit quality but also open new avenues for leveraging cutting-edge LLM advancements. The project's source code is available for community use, promoting collaborative developments in the field\footnote{\href{https://github.com/aixplain/Efficient-MT}{https://github.com/aixplain/Efficient-MT}}. The demo video can be accessed here\footnote{\url{https://youtu.be/vkDH9fC7HfU}}.
\end{abstract}

\section{Introduction}
\begin{figure*}[ht]
\centering
\centerline{\includegraphics[width=\textwidth]{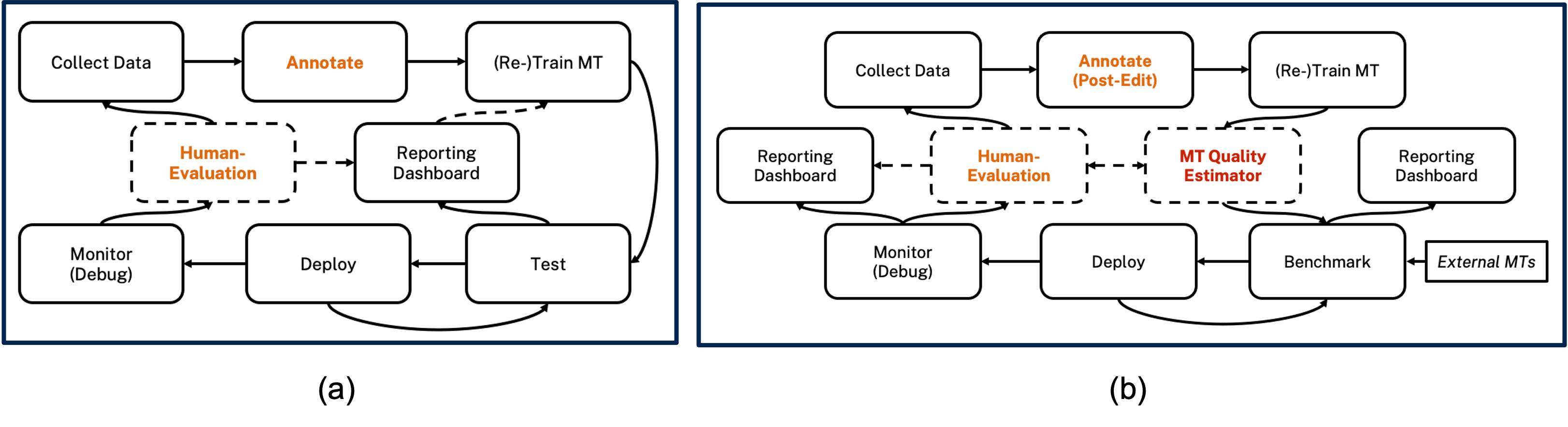}}
\caption{Comparative illustration of MT model evolution. (a) depicts the conventional MT lifecycle, while (b) showcases the enhanced methodology utilizing LLMs for improved efficiency and accuracy.}
\label{fig:MTLifeCycle}
\end{figure*}

Enhancing machine translation (MT) models requires continuous improvements in the accuracy and efficiency of annotation processes. This paper introduces a novel system that integrates large language models (LLMs) with human-in-the-loop annotations to redefine efficiency in MT corpus generation.
Our system, initially focused on human-in-the-loop post-editing for expanding MT corpora, now offers an optional integration with LLM capabilities to refine translation quality and efficiency further. Figure \ref{fig:MTLifeCycle} illustrates the evolutionary progress of our system, highlighting the enhancements and refinements.


Existing systems primarily leverage auto-translation for aiding annotations but often fall short in dynamically integrating the latest advancements in AI, such as real-time learning and adaptive quality estimation based on large language models.
In the previous work \cite{yuksel2023efficient}, a model was developed that enhanced MT models by continuously expanding their corpora through a semi-automated process. The process involved the post-editing of production samples, driven by machine learning models trained by linguists. These models were adept at identifying and prioritizing translations that posed challenges for MT systems, optimizing the post-editing process, and reducing costs. The architecture achieves cost efficiency by selectively targeting translations that challenge MT systems, ensuring human post-editing efforts are concentrated where most impactful. This strategy improves the quality of the MT corpus and reduces costs by decreasing reliance on human input. It also addressed scalability through an automated feedback loop that accommodates growing data volumes without proportionally increasing human labor. 

Central to this approach was the application of active learning techniques, where custom translation quality estimation models were trained in real-time as linguists post-edited. The methodology was supported by insights from the WMT20 and WMT21 Metrics Shared Tasks, which underscored the efficacy of reference-free metrics like COMET-QE in differentiating human translations from MT outputs. The primary contributions of this initial model included a novel architecture for managing the MT model production lifecycle, the strategic use of referenceless metrics for training dataset building and human evaluation, and the active learning of custom referenceless metrics to improve the production lifecycle of MT systems by prioritizing translations for evaluation or post-editing.

Building upon the existing system, significant enhancements were introduced to streamline the MT process further. These advancements include the integration of LLM-generated translations, which mix multiple MT outputs to provide more refined initial hypotheses for post-editing. This approach improves the starting point for linguists and offers a more detailed understanding of translation quality. In addition, the system now employs LLMs for annotation analysis, enabling a more comprehensive evaluation of translation quality. This method leverages the advanced capabilities of LLMs to assess translations with greater depth and accuracy. Another key feature is the incorporation of LLM-based pseudo-labeling. This technique enriches the MT corpus with high-confidence translations, significantly reducing the dependency on human annotation. By automating the identification of reliable translations, the system can efficiently expand its corpus with less manual intervention. Moreover, the system has been equipped to utilize LLMs to suggest the most suitable translation hypothesis. This feature can potentially streamline the translation process by reducing or sometimes even eliminating the need for human annotators in specific contexts. This not only enhances the efficiency of the translation process but also maintains high standards of quality and accuracy.

The primary contributions of this paper are: 
(1) An LLM generates translations by combining multiple machine translation outputs, offering improved initial hypotheses for post-editing.
(2) An LLM provides a more comprehensive evaluation and analysis of annotations, enhancing translation quality assessment.
(3) An LLM is used in pseudo-labeling to enrich the corpus with high-confidence translations, reducing dependence on human annotation.
(4) An LLM suggests the best translation hypothesis, aiding the translation process and potentially replacing human annotators in specific contexts.
This methodology integrates with production systems, using post-edited translations for ongoing MT model improvement. It supports customization and data augmentation, which is vital for MT vendors' personalized services. 

\section{Related Works}
The necessity of research in active learning and quality estimation for MT systems is underscored by the evolving demands of linguistic accuracy and efficiency in translation processes. Quality Evaluation (QE) without reference translations, a concept integral to MT development, primarily focuses on enhancing the quality of MT output during training phases. This aspect of MT is particularly challenging in scenarios where reference translations are unavailable or impractical to generate, a common occurrence in real-world applications.

The inception of QE in the absence of reference translations was marked by the pioneering work of Callison-Burch et al. \cite{callison2012proceedings} at the annual WMT conference. This initiative laid the groundwork for word-level QE, assessing the need for editing translated tokens. The evolution of QE saw the introduction of sentence-level tasks, notably the Direct Assessment method by Graham et al. \cite{graham2015accurate}, which sought to align segment translation scores with human judgment. The WMT-2022 conference further expanded the scope of QE by introducing binary sentence-level classification tasks to identify critical translation errors, as highlighted by Rei et al. \cite{rei2022cometkiwi}.

Recent advancements in QE methodologies have diversified, ranging from the prompt-based learning approach using XLM-R by KU X Upstage \cite{eo2022ku} to the integration of Direct Assessment and MQM features into fine-tuning processes by the Alibaba team \cite{chi2021infoxlm,bao2022alibaba}. Rei et al. \cite{rei2022cometkiwi} introduced an innovative combination of a word-level sentence tagger and explanation extractor within the COMET framework, marking a departure from earlier statistical methods like SVMs, Naive Bayes classifiers, and CRFs \cite{han2013quality}. Kocmi et al. \cite{kocmi-federmann-2023-gemba-mqm,kocmi-federmann-2023-large} proposed an LLM-based translation quality assessment metric, GEMBA, which evaluates the translation of each fragment individually and then averages all the obtained scores to obtain a final system-level score. These developments underscore the dynamic nature of QE in MT, continually adapting to incorporate more sophisticated and varied techniques. 

Parallel to these developments in QE, active learning has emerged as a pivotal strategy for corpus extension in MT. Traditional approaches in this domain have primarily utilized model-free methods based on diversity and model-based uncertainty sampling. Notable works include Peris and Casacuberta \cite{peris2018active}, who leveraged the attention mechanism of neural MT for interactive MT and human supervision, and Zeng et al. \cite{zeng2019empirical}, who employed paraphrastic embeddings from unsupervised pre-training for active learning. Hu and Neubig \cite{hu2021phrase} further contributed to this field by adopting uncertainty-based active learning for fine-tuning MTs, focusing on phrase selection rather than entire sentences.

\section{Methodology}
The methodology of this research incorporates a semi-automated post-editing process augmented by the integration of LLMs to refine the generation of a MT corpus. This process includes the continuous and iterative training of a machine learning model with linguist-provided post-edits and LLM-generated suggestions, enhancing the model's ability to prioritize translations for post-editing and provide real-time feedback.

\subsection{Integration of LLMs and System Enhancements}
The ML model is trained online, using a custom MT quality estimation metric, with a teacher metric (i.e. GEMBA-DA or COMET-QE) as a foundation. These metrics evolve with each post-edit and are further enriched by including LLMs to generate better transitions using multiple translations. By ensembling multiple translations with appropriately designed prompts, the LLMs can synthesize more coherent and contextually appropriate translations. The integration of LLMs allows for a more detailed evaluation of translation quality, leveraging LLMs' vast knowledge and pattern recognition capabilities to inform the quality estimation process.

LLM's role extends to Annotation Analysis, where its evaluation is used to cross-verify the quality of translations and suggest potential improvements or flag errors. This dual-analysis ensures a more comprehensive assessment of each translation's fidelity and naturalness.
Introducing GEMBA as a tool to define translation quality adds a new dimension to quality estimation. GEMBA's advanced understanding of language details aids in creating a more accurate and dynamic quality metric that adapts to the complexities of language and context in real-time.

LLM's pseudo-labeling capability is employed to generate provisional labels for unannotated data, which human annotators can verify or correct. This pseudo-labeling serves as a preliminary filter, reducing the volume of translations requiring human review and focusing efforts on areas where the LLM's confidence is lower.
In scenarios where the quality estimation model suggests high confidence in a translation, LLM's can be proposed to replace human annotators, offering the best translation hypothesis. This suggestion mechanism is designed to support human annotators, providing them with a pre-selected choice that can be approved or further refined, thereby streamlining the post-editing process.

Additionally an ML model has been trained on the fly while the annotations done to estimate the best MT hypothesis and estimate the quality of the translation. The ML model leverages linguistic features and COMET-QE embeddings to estimate translation quality, now complemented by LLM-generated insights. The system \textbf{prioritizes samples} for post-editing based on a combination of the lowest estimated COMET-DA score, the highest estimated Translation-Error-Rate (TER), and LLM evaluations. This tripartite strategy ensures that linguists and LLMs collaborate to effectively enhance the MT model's performance.

\subsection{Technical Implementation}
The technical implementation leverages the FLAML framework for online AutoML functionalities, utilizing fine-tuned cross-lingual embeddings from the COMET-QE score and various linguistic features extracted from source texts and translations. These include token, character, and word length frequencies, as well as Part-of-Speech, Named Entity Recognition, and morphological features. For English sources or target texts, an additional 250 linguistic features are extracted using the LingFeat library in SpaCy.
The AutoML capabilities within the FLAML framework are utilized to fine-tune the selection of features and the training of the quality estimation model. This includes using cross-lingual embeddings and various linguistic indicators that contribute to a robust estimation of translation quality. Additionally, this model aimed to estimate the best MT translation, and using this model, samples were prioritized using uncertainty sampling, further enhancing the efficiency.

\subsection{Evaluation of Translation Quality}
The effectiveness of the enhanced MT system was assessed using various statistical measures and analytical techniques to understand improvements in translation quality and the accuracy of automated estimations.

\begin{itemize}
    \item \textbf{Quality and Character Length Analysis:} The evaluation measured differences in translation quality scores and character lengths before and after post-editing. These metrics highlighted the value added by human annotators and LLMs, and how modifications affect translation length.
    \item \textbf{Correlation and Model Performance Analysis:} Correlation coefficients (Spearman, Pearson, and Kendall) were calculated between COMET-QE metrics and LLM estimations to validate the quality estimations. Additionally, the system's accuracy in identifying top-performing models demonstrated its effectiveness in real-world scenarios, validating the integration of LLM predictions with established metrics like COMET-QE.
    \item \textbf{Error Category Classification:} A multiclass classification automated the identification and analysis of translation errors into categories such as 'Accuracy', 'Terminology', and 'Fluency', among others. This analysis provided a detailed breakdown of precision, recall, and F1-scores, underscoring the system’s capability to handle diverse error types effectively.
\end{itemize}

These evaluations collectively enhance our understanding of the MT system's performance, showcasing the substantial improvements and the system’s ability to adapt to various challenges.


The methodology fosters a continuous improvement cycle for MT models by integrating post-edits back into the training corpus. This cycle improves the quality of the MT outputs and ensures the MT models remain up-to-date and effective in handling new and complex translation tasks. Figure \ref{fig:MTLifeCycle} illustrates the system's dynamic feedback mechanisms.

\section{System Architecture}
Our application is structured to facilitate the MT annotation process, with a dual-sided interface catering to both annotators and administrators. This design allows efficient management of the MT model production lifecycle, enhancing the quality of MT outputs through continuous learning.

\begin{figure}[t]
\centering 
\centerline{\includegraphics[width=\columnwidth]{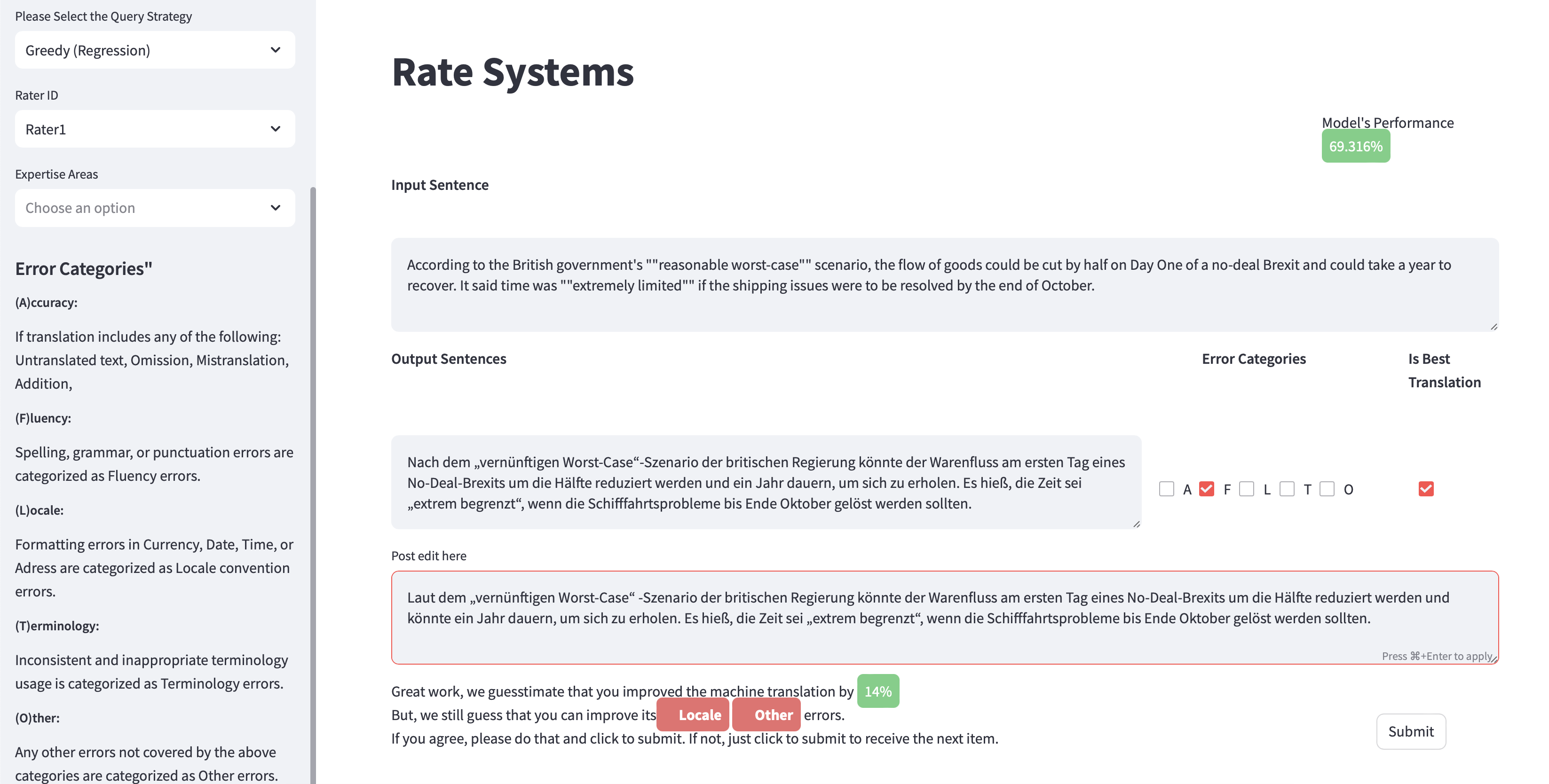}}
\caption{Annotator interface overview. This figure displays the user interface utilized by annotators, detailing the workflow for error categorization, translation selection, and post-editing within the system.}
\label{fig:annotatorUI}
\end{figure}

\subsection{Annotator Interface}
The annotator side of the application, shown in Figure \ref{fig:annotatorUI}, is designed to streamline the post-editing and evaluation of MT outputs. Annotators are presented with sentences translated by MT systems, alongside a set of potential error categories such as accuracy, fluency, local errors, terminology, and others \cite{mqm2022}. The interface is intuitive, providing explanations and examples for each error type to guide the annotator in identifying and categorizing translation issues accurately.

Annotators are equipped with tools to rate translations from multiple MT providers, capable of simultaneously handling up to ten different systems. This multi-rate capability is complemented by a feature that allows annotators to select their expertise areas, ensuring that translations are evaluated by individuals with the most relevant knowledge and experience.
The application also supports an optimization feature that adjusts learning strategies based on the best translation prediction model's performance. This feature ensures that the learning process is robust and tailored to the evolving needs of the MT systems and the nuances of different language pairs.

Upon selecting an error present in a translation, annotators can perform post-edits directly within the app. The system is designed to learn from these post-edits, with models retraining in real-time to incorporate the latest human inputs. This learning mechanism is crucial for improving the MT systems' accuracy and fluency over time.
As annotators select the best translation from the available options, the application prompts them to post-edit the output. These post-edits are then used to refine the MT model further, with the system providing feedback on the improvement percentage and remaining error categories. The LLM-based solution proposes a novel post-editing version, providing detailed feedback on the error categories resolved through the post-editing process. Additionally, it offers real-time feedback to annotators on improving translation quality, measured using the GEMBA metric.

\begin{figure}[t]
\centering 
\centerline{\includegraphics[width=\columnwidth]{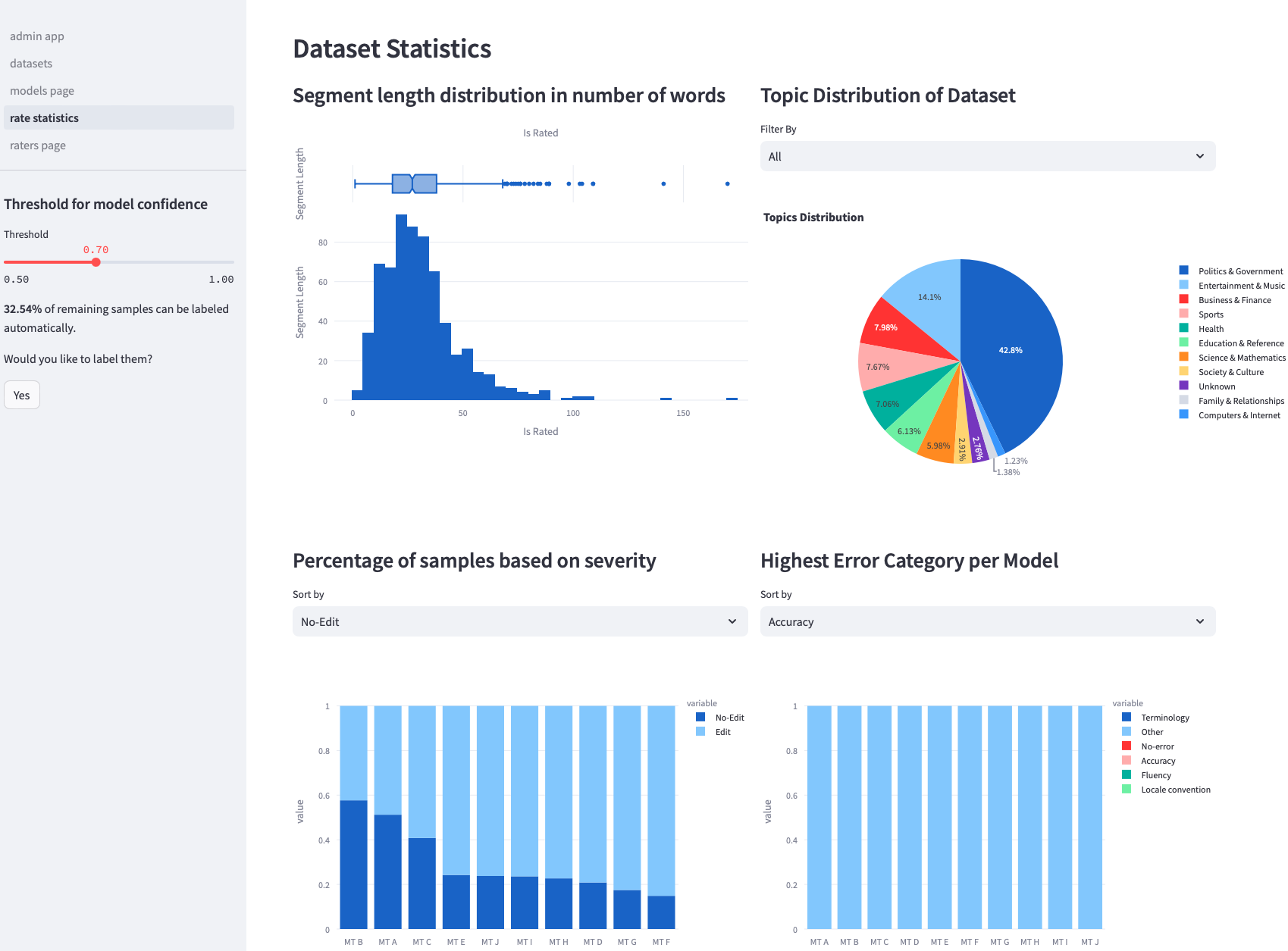}}
\caption{Administrator dashboard visualization: showcasing the admin panel automatic labeling controls, performance metrics across multiple MT systems, comparative analysis of annotator efficiency, and quality estimations.}
\label{fig:adminUI}
\end{figure}

\subsection{Administrator Interface}
The administrator side, shown in Figure \ref{fig:adminUI}, is responsible for overseeing the entire annotation process, managing annotator assignments, and ensuring the quality of the MT models. It provides a comprehensive dashboard for tracking progress, analyzing error trends, and making informed decisions about the MT systems' deployment and retraining schedules.
The application offers real-time statistics on the number of samples rated and the performance of the MT models based on the annotators' inputs. This feature is vital for administrators to monitor the annotation process's effectiveness and the MT outputs' overall quality.

At the forefront of the admin dashboard is the model confidence feature, which displays the percentage of samples that can be automatically labeled by the system's models. Administrators can set a confidence threshold to determine the extent of automatic labeling versus human annotation. This adjustable threshold allows for flexibility in balancing between automated processes and human oversight. The interface prompts administrators with the option to label samples automatically based on this threshold, showcasing the system's capability to reduce manual labeling efforts progressively.

A critical function of the admin interface is the ability to analyze each MT system's performance based on human-labeled data. Adding a 'no edit' category allows administrators to quickly identify which MT outputs are deemed satisfactory without any required edits, thus gauging the effectiveness of different MT systems. Detailed error categorization per model is also available, providing insights into specific areas where each MT system may excel or underperform.

By leveraging the data collected from annotators and the system's predictive models, administrators can make informed decisions on MT system improvements and annotator training needs. The interface allows for evaluating unreviewed samples, predicting which MT system might perform best based on historical data and current model training.
The admin app replicates benchmarking reports, serving as a quality control mechanism by highlighting the performance of MT systems against human standards. It enables administrators to thoroughly review the systems, identify trends, and implement strategic improvements.

The admin app visualizes various data points, such as segment length and classified topics within the dataset. These visualizations aid in understanding the distribution and characteristics of the data being processed. Administrators can filter these views based on whether the data has been rated or not, enabling a focused analysis of specific subsets of the dataset.

The admin interface provides tools for evaluating annotator performance, allowing administrators to view and compare statistics on the number of samples processed by each annotator. This comparison helps identify potential biases or discrepancies in error category selection. A key feature of the enhanced system is the correlation analysis between LLM estimations and human annotator labels, which is essential for monitoring performance quality. By aligning LLM estimations with human judgments, the system can promptly identify and address any inconsistencies or declines in performance, thereby enhancing the reliability and accuracy of the translation process.

\section{Results}

This section presents the results of the evaluation, demonstrating the effectiveness of the enhanced machine translation system integrated with large language models (LLMs). The analysis included a variety of metrics to measure translation quality improvements, model performance, and the accuracy of the quality estimation methods.

\subsection{Translation Quality Improvement}

The system showed a significant improvement in translation quality post-editing. The average improvement in translation quality scores was $4.33\%$, with a standard deviation of $10.25\%$, indicating a consistent enhancement across various translations. Additionally, the absolute difference in character count between post-edited translations and the original target translations averaged $16.35$, suggesting that post-editing not only enhances quality but also adjusts translation length to better match the source content.

\subsection{Correlation Analysis}

The correlation analysis between COMET-QE and automated quality estimates with LLMs was performed using Spearman correlation coefficients. The Spearman coefficient was $0.40$, indicating a moderate correlation, and significant alignment with COMET-QE and validating the reliability of the automated quality metrics in reflecting human judgment.

\subsection{Model Performance and Accuracy}

The system's capability to identify the best-performing translation model for each segment was evaluated. 
\begin{table}[ht]
\centering
\caption{Accuracy Scores for Best Model Prediction}
\label{tab:weighted_avg}
\begin{tabular}{lcc}
\hline
\textbf{Model} & \textbf{Top-1 Acc.} & \textbf{Top-3 Acc.} \\ \hline\hline
LLM-Based & $24\%$ & $57\%$ \\ \hline
Online Model & $12\%$ & $25\%$ \\ \hline
\end{tabular}
\end{table}

The analysis indicated that the system could accurately select the best model with an accuracy of $24$\% for the top-1 model and $57$\% for the top-3 models. On the other hand, Ranking model that is trained to estimate the best model on the fly has an accuracy of $12\%$ for top-1 and $25\%$ for top-3 models. This demonstrates the ability of LLM based model selection in a competitive translation environment.





These results underscore the effectiveness of integrating LLMs into the MT annotation process, enhancing the overall quality and efficiency of the machine translation output. The continuous improvement cycle facilitated by this system ensures that MT models are dynamically updated, maintaining high translation quality across various languages and contexts.

\section{Conclusion}
\label{sec:conclusion}
Integrating LLMs into the MT corpus generation has enhanced the efficiency and effectiveness of translation processes. This study demonstrated that augmenting a semi-automated post-editing framework with LLM capabilities significantly improves system functionality, offering more accurate translation quality assessments and reducing the reliance on extensive human input. The use of LLMs for annotation analysis and pseudo labeling optimizes the post-editing process by providing real-time, quality-driven feedback and suggestions, reducing the cognitive load on annotators and enabling them to concentrate on complex tasks. These advancements not only streamline the annotation workflow but also promote a competitive and engaging work environment. The adaptability of this system accommodates various operational needs, ensuring it remains effective across different linguistic contexts.

\section{Ethics and Broader Impact}
The ethical and broader impact considerations highlighted here underscore the importance of responsible development and deployment of the proposed MT corpus generation methodology. By addressing these considerations, we aim to ensure that the proposed efficient MT corpus generation process contributes positively. It is essential to remain vigilant, iterate on the system with ethical principles in mind, and engage with stakeholders to gather diverse perspectives for ongoing improvement.

The quality estimation process should be conducted with vigilance to mitigate biases in the training data. LLMs may inadvertently perpetuate or amplify existing biases in the data they are trained on. It is crucial to regularly audit and address bias in the training data and the LLM-generated translations to ensure fairness and avoid reinforcing stereotypes or marginalizing certain language groups.

For human post-editing, adhering to strict data privacy protocols and obtaining informed consent from users contributing to the training data is imperative. Additionally, sensitive or personal information should be handled carefully to protect user privacy. The increased automation introduced by LLMs in the translation process may impact human annotators. It is crucial to consider the potential displacement of jobs and provide training opportunities for annotators to adapt to new roles or responsibilities within the evolving MT ecosystem.

\bibliography{main}

\begin{thebibliography}{14}
\providecommand{\natexlab}[1]{#1}

\bibitem[{Bao et~al.(2022)Bao, Wan, Liu, Yang, Lei, He, Wong, and Xie}]{bao2022alibaba}
Keqin Bao, Yu~Wan, Dayiheng Liu, Baosong Yang, Wenqiang Lei, Xiangnan He, Derek~F. Wong, and Jun Xie. 2022.
\newblock Alibaba-translate china’s submission for wmt 2022 quality estimation shared task.
\newblock In \emph{Proceedings of the Seventh Conference on Machine Translation (WMT)}, pages 597--605, Abu Dhabi, United Arab Emirates (Hybrid). Association for Computational Linguistics.

\bibitem[{Callison-Burch et~al.(2012)Callison-Burch, Koehn, Monz, Post, Soricut, and Specia}]{callison2012proceedings}
Chris Callison-Burch, Philipp Koehn, Christof Monz, Matt Post, Radu Soricut, and Lucia Specia. 2012.
\newblock Proceedings of the seventh workshop on statistical machine translation.
\newblock In \emph{Proceedings of the Seventh Workshop on Statistical Machine Translation}, Montreal, Canada. Association for Computational Linguistics.

\bibitem[{Chi et~al.(2021)Chi, Dong, Wei, Yang, Singhal, Wang, Song, Mao, Huang, and Zhou}]{chi2021infoxlm}
Zewen Chi, Li~Dong, Furu Wei, Nan Yang, Saksham Singhal, Wenhui Wang, Xia Song, Xian-Ling Mao, Heyan Huang, and Ming Zhou. 2021.
\newblock Infoxlm: An information-theoretic framework for cross-lingual language model pre-training.
\newblock In \emph{Proceedings of the 2021 Conference of the North American Chapter of the Association for Computational Linguistics: Human Language Technologies}, pages 3576--3588, Online. Association for Computational Linguistics.

\bibitem[{Eo et~al.(2022)Eo, Park, Moon, Seo, and Lim}]{eo2022ku}
Sugyeong Eo, Chanjun Park, Hyeonseok Moon, Jaehyung Seo, and Heui-Seok Lim. 2022.
\newblock Ku x upstage’s submission for the wmt22 quality estimation: Critical error detection shared task.
\newblock In \emph{Proceedings of the Seventh Conference on Machine Translation (WMT)}, pages 606--614.

\bibitem[{Graham et~al.(2015)Graham, Baldwin, and Mathur}]{graham2015accurate}
Yvette Graham, Timothy Baldwin, and Nitika Mathur. 2015.
\newblock Accurate evaluation of segment-level machine translation metrics.
\newblock In \emph{Proceedings of the 2015 Conference of the North American Chapter of the Association for Computational Linguistics: Human Language Technologies}, pages 1183--1191.

\bibitem[{Han et~al.(2013)Han, Lu, Wong, Chao, He, and Xing}]{han2013quality}
Aaron Li-Feng Han, Yi~Lu, Derek~F Wong, Lidia~S Chao, Liangye He, and Junwen Xing. 2013.
\newblock Quality estimation for machine translation using the joint method of evaluation criteria and statistical modeling.
\newblock In \emph{Proceedings of the Eighth Workshop on Statistical Machine Translation}, pages 365--372.

\bibitem[{Hu and Neubig(2021)}]{hu2021phrase}
Junjie Hu and Graham Neubig. 2021.
\newblock Phrase-level active learning for neural machine translation.
\newblock \emph{arXiv preprint arXiv:2106.11375}.

\bibitem[{Kocmi and Federmann(2023{\natexlab{a}})}]{kocmi-federmann-2023-gemba-mqm}
Tom Kocmi and Christian Federmann. 2023{\natexlab{a}}.
\newblock Gemba-mqm: Detecting translation quality error spans with gpt-4.
\newblock In \emph{Proceedings of the Eighth Conference on Machine Translation}, Singapore. Association for Computational Linguistics.

\bibitem[{Kocmi and Federmann(2023{\natexlab{b}})}]{kocmi-federmann-2023-large}
Tom Kocmi and Christian Federmann. 2023{\natexlab{b}}.
\newblock \href {https://aclanthology.org/2023.eamt-1.19} {Large language models are state-of-the-art evaluators of translation quality}.
\newblock In \emph{Proceedings of the 24th Annual Conference of the European Association for Machine Translation}, pages 193--203, Tampere, Finland. European Association for Machine Translation.

\bibitem[{{MQM Typology}(2022)}]{mqm2022}
{MQM Typology}. 2022.
\newblock \href {https://themqm.org/the-mqm-typology/} {{Multidimensional Quality Metrics (MQM)}}.

\bibitem[{Peris and Casacuberta(2018)}]{peris2018active}
{\'A}lvaro Peris and Francisco Casacuberta. 2018.
\newblock Active learning for interactive neural machine translation of data streams.
\newblock \emph{arXiv preprint arXiv:1807.11243}.

\bibitem[{Rei et~al.(2022)Rei, Treviso, Guerreiro, Zerva, Farinha, Maroti, De~Souza, Glushkova, Alves, Lavie et~al.}]{rei2022cometkiwi}
Ricardo Rei, Marcos Treviso, Nuno~M Guerreiro, Chrysoula Zerva, Ana~C Farinha, Christine Maroti, Jos{\'e}~GC De~Souza, Taisiya Glushkova, Duarte~M Alves, Alon Lavie, et~al. 2022.
\newblock Cometkiwi: Ist-unbabel 2022 submission for the quality estimation shared task.
\newblock \emph{arXiv preprint arXiv:2209.06243}.

\bibitem[{Yuksel et~al.(2023)Yuksel, Gunduz, Sharma, and Sawaf}]{yuksel2023efficient}
Kamer~Ali Yuksel, Ahmet Gunduz, Shreyas Sharma, and Hassan Sawaf. 2023.
\newblock Efficient machine translation corpus generation.
\newblock \emph{arXiv preprint arXiv:2306.11838}.

\bibitem[{Zeng et~al.(2019)Zeng, Garg, Chatterjee, Nallasamy, and Paulik}]{zeng2019empirical}
Xiangkai Zeng, Sarthak Garg, Rajen Chatterjee, Udhyakumar Nallasamy, and Matthias Paulik. 2019.
\newblock Empirical evaluation of active learning techniques for neural mt.
\newblock In \emph{Proceedings of the 2nd Workshop on Deep Learning Approaches for Low-Resource NLP (DeepLo 2019)}, pages 84--93.

\end{thebibliography}

\end{document}